\documentclass[letterpaper, 10 pt, conference]{ieeeconf}  % Comment this line out if you need a4paper

\IEEEoverridecommandlockouts                              % This command is only needed if 
\usepackage{graphicx}
\usepackage[table]{xcolor}
\usepackage{tcolorbox}
\usepackage{hhline}
\usepackage{hyperref}
\usepackage[margin=0.8in]{geometry}

\title{\LARGE Fast Region Proposal Learning for Object Detection for Robotics
}

\author{Federico Ceola$^{1,2}$, Elisa Maiettini$^{1}$, Giulia Pasquale$^{1}$, Lorenzo Rosasco$^{2,3}$ and Lorenzo Natale$^{1}$% <-this % stops a space
\thanks{$^{1}$Federico Ceola, Elisa Maiettini, Giulia Pasquale and Lorenzo Natale are with Humanoid Sensing and Perception, Istituto Italiano di Tecnologia, Genoa, Italy
        {\tt\footnotesize name.surname@iit.it}}
\thanks{$^{2}$Lorenzo Rosasco and Federico Ceola are with Laboratory for Computational and Statistical Learning and with Dipartimento di Informatica, Bioingegneria, Robotica e Ingegneria dei Sistemi, University of Genoa, Genoa, Italy}
\thanks{$^{3}$Lorenzo Rosasco is also with Istituto Italiano di Tecnologia and Massachusetts Institute of Technology, Cambridge, MA {\tt\footnotesize lrosasco@mit.edu}}
}

\begin{document}
\maketitle
\thispagestyle{empty}
\pagestyle{empty}

%===============================================================================

\begin{abstract}
Object detection is a fundamental task for robots to operate in unstructured environments. 
Today, there are several deep learning algorithms that solve this task with remarkable performance. Unfortunately, training such systems requires several hours of GPU time. 
For robots, to successfully adapt to changes in the environment or learning new objects, it is also important that object detectors can be re-trained in a short amount of time. 
A recent method~\cite{maiettini2019a} proposes an architecture that leverages on the powerful representation of deep learning descriptors, while permitting fast adaptation time. Leveraging on the natural decomposition of the task in (i) regions candidate generation, (ii) feature extraction and (iii) regions classification, this method performs fast adaptation of the detector, by only re-training the classification layer. This shortens training time while maintaining state-of-the-art performance. 
In this paper, we firstly demonstrate that a further boost in accuracy can be obtained by adapting, in addition, the regions candidate generation on the task at hand. Secondly, we extend the object detection system presented in~\cite{maiettini2019a} with the proposed fast learning approach, showing experimental evidence on the improvement provided in terms of speed and accuracy on two different robotics datasets. The code to reproduce the experiments is publicly available on GitHub\footnote{\url{https://github.com/robotology/online-detection}}.
\end{abstract}

\begin{keywords}
%List of keywords (from the RA Letters keyword list)
% TODO change this
%Object Detection, Fast Domain Adaptation, Kernel Methods
Visual Learning, Object Detection, Segmentation and Categorization
\end{keywords}

% Two or three meaningful keywords should be added here
%\keywords{Object Detection, Fast Domain Adaptation, Kernel Methods} 

%===============================================================================

%%%%%%%%%%%%%%%%%%%%%%%%%%%%%%%%%%%%%%%%%%%%%%%%%%%%%%%%%%%%%%%%%%%%%%%%%%%%%%%%
\section{INTRODUCTION}
\label{sec:introduction}
%!TEX root = ../root.tex

High reliability and quick adaptation are two fundamental features for several robotic perception systems. Latest advancements in the deep learning brought stunning improvements in terms of accuracy and precision in many visual perceptive tasks like e.g. object detection and segmentation~\cite{He2015,ren2015_faster,He2017,yolov4}. Such performance are appealing in robotics, however, the large amount of parameters in deep learning architectures, makes their optimization process highly demanding in terms of annotated data and training time.
Moreover, some visual tasks, like object detection or segmentation, have inherent complexity (e.g., the large and unbalanced set of associated examples) which further increases the computational resources required at training time.
%
%increments the computational resources required by methods that address them, dramatically slowing down their adaptation capabilities to new tasks.\\
In this work, we focus on the problem of object detection, which is at the basis of more sophisticated tasks in robotics like segmentation and pose estimation. A recently proposed approach~\cite{maiettini2019a} presented a fast learning strategy that allows to adapt an object detection system to a new task in a fraction of the training time of the conventional methods, while maintaining state-of-the-art accuracy. This method leverages on the natural decomposition of the object detection task in three fundamental steps --namely (i) region candidates prediction on the target image, (ii) per region feature extraction and (iii) regions classification-- to speed up the training of the detection pipeline. Specifically, \cite{maiettini2019a} proposes a two-stages method composed of: (i) a region and feature extractor, trained once, off-line, on a certain task, based on Faster R-CNN's~\cite{ren2015_faster} first layers, followed by (ii) a region classifier which instead is designed to be trained quickly, on-line, on the target task. In this paper, we demonstrate that, while efficient, the lack of adaptation of the region and feature extractor is a limitation that leads to a drop in performance~\cite{maiettini2019a}, that can be partially or even completely recovered by also adapting the region proposal layers on the target task. We report on empirical evidences of this statement in Sec.~\ref{experiments:preliminary}. Motivated by this analysis, we developed a new architecture and training strategy that adapts both the region proposal and the classification layers in few minutes. The experimental analysis shows that we effectively achieve higher performance than the work in~\cite{maiettini2019a}, and that our training method significantly reduces training time with respect to a baseline obtained by a conventional fine-tuning of a deep neural architecture. Additionally, we show that the proposed hybrid method has better properties against over-fitting at training time than the chosen deep learning based baseline. To summarize, the contribution of this work are as follows:
\begin{enumerate}
	\item We show empirical evidences of the importance of region proposal network's adaptation on the task at hand.
	\item We design a fast learning approach to train a region proposal method, by adapting an on-line learning strategy~\cite{maiettini2019a} to the region candidates generation.
	\item We propose a novel efficient object detection pipeline, building on~\cite{maiettini2019a}. Firstly, we improve the feature extractor in~\cite{maiettini2019a} by substituting Faster R-CNN's~\cite{ren2015_faster} first layers with Mask R-CNN's~\cite{He2017} first layers. Secondly and more importantly, we substitute the region proposal network of the feature extractor with the proposed fast learning approach for region proposal generation.
	\item We report on an empirical evaluation of the resulting pipeline on two robotics benchmarks
	and we carry out an experimental analysis on the variation of one of its main parameters.
\end{enumerate}

The remaining of the paper is organized as follows: in Sec.~\ref{sec:relwork}, we overview main state-of-the-art methods for object detection in robotics, focusing on region proposal generation. In Sec.~\ref{sec:methods}, we describe the proposed pipeline, focusing on the fast learning of the region proposal network. Then, in Sec.~\ref{sec:experiments}, we benchmark the pipeline on two different robotics datasets and empirically analyze it. Finally, in Sec.~\ref{sec:conclusions}, we discuss the obtained results and conclude drawing lines for future work.

%%%%%%%%%%%%%%%%%%%%%%%%%%%%%%%%%%%%%%%%%%%%%%%%%%%%%%%%%%%%%%%%%%%%%%%%%%%%%%%%
\section{RELATED WORK}
\label{sec:relwork}
%!TEX root = ../root.tex
%In this section we provide an overview of the state-of-the-art approaches for object detection, focusing on the robotic requirement of fast adaptation and on an overview of different localization techniques.
%While discussing the state-of-the-art, we illustrate our contribution.
%\subsection{Object Detection for robotics}
%\label{relwork:detection}
%Object Detection is the task of localizing, by means of bounding boxes, and recognizing objects of interest in 2D images~\cite{ODsurvey1,ODsurvey2}. 
In this section, we overview state-of-the-art approaches for object detection, focusing on the localization problem and the robotic requirement of fast adaptation, which is the main motivation for our contribution.
\subsection{Object Detection for robotics}
\label{relwork:detection}
Three main steps can be identified to accomplish the object detection task~\cite{ODsurvey1,ODsurvey2}, i.e. (i) region candidates generation, (ii) per region feature extraction and (iii) region classification and refinement. The major trend in the state-of-the-art is to integrate the three steps into ``monolithic'' deep learning based architectures~\cite{He2017,yolov4,zhai2020,duan2019,rua2020,Tan2020}. These methods achieve remarkable performance on the main object detection benchmarks~\cite{pascal2010,coco}, at the cost of long training time (hours or even days of training), mainly due to the large amount of parameters that need to be optimized on the data. Such training time is problematic for robotics, because it prevents on-line adaptation.\\
%The great reliability in terms of accuracy of these approaches is appealing for vision systems for robotics, however the capacity of quickly adapt to new tasks is fundamental for robots that operate in dynamical scenarios. 
Latest literature on object detection for robotics either focuses on improving robustness and precision in particular scenarios (like occlusions in clutter~\cite{Zhang2020,Novkovic2020}) or on finding solutions to obtain state-of-the-art accuracy, while allowing for fast adaptation to new tasks~\cite{maiettini2018,maiettini2019a}. For instance, the solution proposed in~\cite{maiettini2019a} is an hybrid architecture that integrates a deep learning approach for region and feature extraction~\cite{ren2015_faster} with an optimized kernel based method~\cite{falkon2018,falkonlibrary2020} for region classification. The key aspect in~\cite{maiettini2019a} is the decoupling of the learning of the feature extractor, based on a set of convolutional layers and a Region Proposal Network (RPN)~\cite{He2017}, and the optimization of the detector, based on kernel methods~\cite{falkonlibrary2020} and on an approximated approach for negatives selection, called Minibootstrap~\cite{maiettini2019a}. The former is performed only once and off-line on the available data, while the latter is performed on-line on the task at hand, when new data becomes accessible.\\
The approach proposed in~\cite{maiettini2019a} allows to quickly re-train a detector in presence of new objects, with a convenient speed/accuracy trade-off. In this work, we demonstrate that an adaptation of the region proposal method is also important to increase performance. Thus, we present a novel pipeline, building on previous work~\cite{maiettini2019a} which allows to adapt both the RPN and the classification layers on the task at hand, in a fraction of the time required by a conventional fine-tuning approach, further pushing the aforementioned speed/accuracy trade-off. 

%\begin{enumerate}
%%	\item Introduction and definition of the task~\cite{ODsurvey1,ODsurvey2}
%%	\item Main methods of the SoA~\cite{girshick15_fastrcnn,ren2015_faster,He2017,redmon2016}
%%	\item State of the art results are stunning and represent a great opportunity for robotics, however deep learning based methods are really slow to adapt and this is not good for robotics.
%%	\item For this reason, solutions has been proposed~\cite{maiettini2018,maiettini2019a}, which proposing a hybrid solution, integrating cnn and kernel based methods manage to leverage on both (i) powerful representation of deep networks and (ii) fast adaptation capabilities of recently proposed kernel based methods, specifically optimized for large scale datasets~\cite{falkon2018}.
%%	\item Key aspect of these approaches is the decoupling of the learning of the feature extractor, done once and offline on the available data, and the optimization of the detector, based on kernel methods, on the task at hand, done online in few seconds.
%	\item However, even if effective and efficient for the fast adaptation, lack of adaptation of the feature extractor on the task at hand, might lead to gap in accuracy, especially when the task at hand is very different from the feature task.
%	\item For this reason, in this work, we propose an evolution of the online learning pipeline proposed in~\cite{maiettini2018,maiettini2019a}
%\end{enumerate}
%
%\subsection{Localization methods for object detection}
%\label{relwork:localization}
\subsection{Localization methods for object detection}
\label{relwork:localization}
A na\"ive approach to detect objects in an image is that of visiting \textit{all} the possible locations in a sliding window fashion and classifying them to discard background areas. This approach was extensively used~\cite{Viola2001} but it is computationally heavy. Lately, new object detection methods have been devised to address the localization step, shrinking the list of regions to consider for classification.\\
\noindent{{\bf Region-based approaches.}} Region-based object detection methods~\cite{ren2015_faster,dai2016,He2017} use the so called \textit{region proposal} approaches to produce a reduced set of \textit{Regions of Interest} (RoIs) to be classified. These methods can be either general purpose, based on geometrical properties of the image, to e.g. group \textit{similar} pixels~\cite{uijlings2013}, or they can be optimized, i.e. learned, on the task at hand~\cite{ren2015_faster}. For instance, the \textit{Region Proposal Network} (RPN)~\cite{ren2015_faster} is a 2-layers fully convolutional network which predicts region candidates on the convolutional feature map of the input image. This process is based on the concept of \textit{anchors} (introduced in~\cite{ren2015_faster} and widely used in the subsequent literature~\cite{dai2016,He2017,yolov4}) to encode different scales and aspects ratio of a single location on the feature map. After RPN introduction~\cite{ren2015_faster}, a great work has been done in order to improve it in terms of precision of the predicted regions~\cite{hosang2016,vu2019,galteri2017}. Anchors introduction shapes the problem of localization problem from a 4-coordinates prediction to a 4-offsets regression.\\
\noindent{{\bf Grid-based approaches.}} These methods are based on anchors~\cite{yolov3,yolov4,zhai2020} as well (also called \textit{default boxes}). These approaches remove the intermediate step of region candidates proposal, directly producing classification predictions on the different anchors. While being typically more efficient in terms of training and inference time, they are also slightly less precise on standard benchmarks than the region-based methods~\cite{ODsurvey1}.\\
\noindent{{\bf Keypoint based approaches.}} Differently from the above two approaches, these methods~\cite{law2018,duan2019,zhou2019,rua2020} typically represent objects of interest as a set of keypoints (e.g. the bounding box's center~\cite{zhou2019} or top-left and bottom-right corners~\cite{law2018}). The keypoints are predicted by e.g. fully convolutional networks~\cite{law2018,zhou2019} and encoded into an embedding which is used for classification and objects properties regression.\\

All the aforementioned methods for localization, based on learning techniques, are typically \textit{embedded} in the detection network and are optimized on the task at hand via end-to-end backpropagation of the whole network, requiring long training time. In this work, we propose, a fast training method for both the classification and the RPN, demonstrating higher performance than previous work and much shorter training time than a conventional fine tuning protocol.

%\begin{enumerate}
%%	\item Grid based approaches~\cite{redmon2016,redmon2016yolo9000,liu2015_ssd}
%%	\item keypoint detection~\cite{duan2019} objects as points~\cite{zhou2019}, cornernet~\cite{law2018}
%%	\item Region based approach~\cite{girshick2014_rcnn,girshick15_fastrcnn,ren2015_faster,He2017,dai2016}
%
%%	RPN analysis~\cite{hosang2016}, cascade RPN~\cite{vu2019}, adaptive RPN~\cite{lu2019}
%	\item Our solutions
%\end{enumerate}

%\subsection{Few shots learning for Object Detection (?)}
%\label{relwork:fewshots}
%
%\subsection{Weak Supervision}
%\label{relwork:weaksupervision}
%\begin{enumerate}
%	\item another way to address the generalization problem is by using weak supervision
%	\item however having good features/regions extractor for the task at hand is fundamental
%\end{enumerate}

%%%%%%%%%%%%%%%%%%%%%%%%%%%%%%%%%%%%%%%%%%%%%%%%%%%%%%%%%%%%%%%%%%%%%%%%%%%%%%%%
\section{METHODS}
\label{sec:methods}
%!TEX root = ../root.tex
\begin{figure*}
    %\vspace*{10mm}
	\centering
	\includegraphics[width=0.65\linewidth]{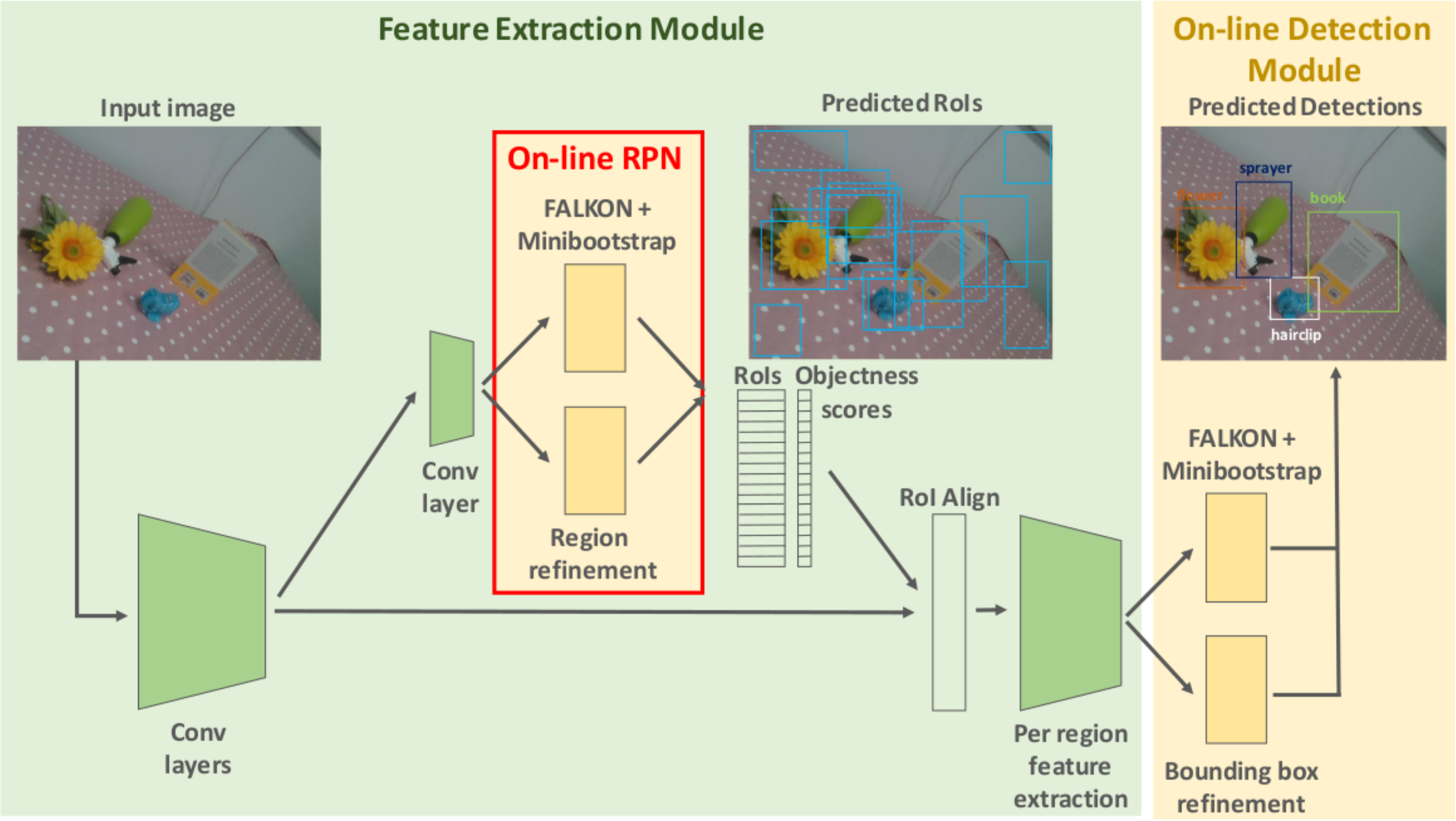}
	\caption{Overview of the proposed object detection pipeline. The \textit{Feature Extraction Module} relies on Mask R-CNN architecture and the proposed On-line RPN, to extract deep features and predict RoIs from each input image. The \textit{On-line Detection Module} performs RoIs classification and refinement, providing as output the detections for the input image. The green blocks are trained off-line on the FEATURE-TASK, while the yellow blocks are trained on-line on the TARGET-TASK. For a detailed description of the pipeline, see Sec.~\ref{methods:overview}.}
	\label{fig:rpnpipeline}
\end{figure*}
We consider the scenario in which a robot is asked to learn to detect a set of novel object instances (referred to as TARGET-TASK) during a brief interaction with a human with a teacher-learner pipeline, as in~\cite{maiettini2017}. The detection system is trained on-line on the TARGET-TASK, by exploiting some components of the pipeline previously trained on a different task (referred to as FEATURE-TASK).\\
In this section, we describe the proposed detection method. Specifically, in Sec.~\ref{methods:overview}, we overview the proposed system, while in Sec.~\ref{methods:onlinelearning}, we focus on the main contribution of this paper, i.e., the strategy that we apply to learn on-line the region proposals on the TARGET-TASK.

\subsection{Overview of the pipeline}
\label{methods:overview}
The proposed detection pipeline (see Fig.~\ref{fig:rpnpipeline}) is composed of two main modules: (i) a \textit{Feature Extraction Module} and (ii) an \textit{On-line Detection Module}. For the former one, which computes a convolutional feature map for each proposed RoI, we rely on the first layers of Mask R-CNN where we replace the last sibling layers of the RPN with the proposed fast learning region proposal method (namely, On-line RPN in Fig.~\ref{fig:rpnpipeline}). For the \textit{On-line Detection Module}, which uses the provided per RoI convolutional features for classification and refinement, we replace the two output layers of Mask R-CNN for class prediction and bounding box regression, with the on-line object detection method~\cite{maiettini2019a}.\\
The main contribution of the proposed approach relies on the \textit{Feature Extraction Module}. Firstly, with respect to~\cite{maiettini2019a}, we substitute the first layers of Faster R-CNN~\cite{ren2015_faster}, with the first layers of Mask R-CNN~\cite{He2017}, allowing to increase the quality of the extracted feature. Indeed, the adoption of the \textit{RoI Align} layer~\cite{He2017} allows to reduce the misalignment in the feature extraction process given by the coarse spatial quantization previously performed in the \textit{RoI Pooling} layer~\cite{ren2015_faster}. Secondly and more importantly, we propose a method for quickly re-training a region proposal network, the \textit{On-line RPN}. More details of this latter contribution are provided in Sec.~\ref{methods:onlinelearning}.\\
\noindent{{\bf Learning protocol.}} The learning process of the pipeline can be divided in two stages. The off-line stage is performed by training Mask R-CNN for object detection on the FEATURE-TASK, following the training procedure proposed in~\cite{He2017}. The weights learned during this off-line stage, represented by the green blocks in Fig.~\ref{fig:rpnpipeline} (the convolutional layers) are then used in the second learning stage.\\
The on-line learning stage is performed on the TARGET-TASK and it can be divided in two sub-steps: firstly the \textit{On-line RPN} is optimized using the features extracted by the shared convolutional feature extractor previously trained, subsequently, the \textit{On-line Detection Module} is trained, by using the per region features proposed by the \textit{On-line RPN} jointly with the following convolutional layers. While the first off-line stage is performed only once on the available data, this second on-line stage is performed in few minutes on the task at hand.

% spatially aggregate the activations of the convolutional feature map of an image within the area defined by each proposed RoI.
%Specifically, for the classification of the proposed RoIs, we design a novel method which employs a set of FALKON binary classifiers; for the bounding box refinement, instead, we simply rely on the Regularized Least Squares (RLS) regression approach proposed in Region-CNN (Girshick et al., 2014).
%\begin{enumerate}
%	\item List all the components of the pipeline
%	\item Describe how each model is connected to each other
%%	\item Describe the choices for the backbone
%\end{enumerate}
\subsection{On-line learning strategy}
\label{methods:onlinelearning}
In order to quickly learn region candidates for the task at hand, we propose an adaptation of the strategy from~\cite{maiettini2019a}. In that case, the output layers of Faster R-CNN were replaced, respectively with an optimized Kernel method (FALKON)~\cite{falkon2018,falkonlibrary2020} for classification, and a Regularized Least Squares (RLS) regressor for bounding box refinement~\cite{girshick2014_rcnn} (\textit{On-line Detection Module} in Fig.\ref{fig:rpnpipeline}).
%Respectively, the former one was substituted with a set of FALKON binary classifiers~\cite{falkon2018}, while for the latter one, the Regularized Least Squares (RLS) regression approach proposed in Region-CNN~\cite{girshick2014_rcnn} was considered.\\
In addition,~\cite{maiettini2019a} proposed an approximated \textit{Hard Negative Mining method}~\cite{Sung1996,Felzenszwalb2010,girshick2014_rcnn}, called Minibootstrap, to efficiently select the hard negatives from the dataset, addressing the well known problem of positives-negatives imbalance in object detection~\cite{Lin2017focal}.\\
In this work, we propose to use this learning strategy for both the \textit{On-line Detection Module} and the region proposal method. In the following paragraph, we illustrate how we propose to adapt this strategy to obtain the \textit{On-line RPN} (see Fig.~\ref{fig:rpnpipeline}). Our motivation is that the region proposal task is similar to the region classification performed by the \textit{On-line Detection Module}, but for the definition of classes. Specifically, in the region proposal step, there is a single foreground class, comprising all target classes altogether, while the background class is strongly redundant. In the region classification step, instead, a specific class for every target is defined (since the objects have to be identified), while the background class is composed by all the wrong proposed regions (usually this class is still dominant, but has already went through the region proposal selection process).\\

\noindent{{\bf On-line RPN.}}
While in principle the RPN's problem is similar to the detection one (i.e. region classification and refinement), in~\cite{He2017} they are addressed differently. The RPN is an anchor based localization method, i.e. it uses different anchors (representing different scales and aspect ratios) to predict the objectness of the regions in an image (see~\cite{ren2015_faster} for further details). Given a convolutional feature map of size $h \times w \times f$ (where $h$ and $w$ are the height and width of the feature map and $f$ is the number of channels), the per anchor region classifier in the RPN in~\cite{ren2015_faster} is implemented as a convolutional layer with $A$ channels, kernel size 1 and stride 1 (where $A$ is the number of anchors). The output tensor is, hence, of size $h \times w \times A$, where the element $ija$ represents the objectness score of the location $ij$ for the $a^{th}$ anchor. All the regions with an Intersection over Union (IoU) greater than 0.7 with the ground truth are used as positives (foreground) (alternatively, if a ground truth does not have regions with an IoU greater than 0.7, the ones with the highest IoU are considered), while all the regions with an IoU smaller than 0.3 are considered as negatives (background).\\
In this work, we substitute the $A$ kernels for classification with $A$ FALKON binary classifiers. We ``unroll'' the $h \times w \times f$ feature map into a list of $hw$ features of size $f$. All of them represent specific locations in the original image and they are used as learning samples by the classifiers, trained within the Minibootstrap procedure~\cite{maiettini2019a}. 
As explained in~\cite{maiettini2019a}, the Minibootstrap visits a random subset of negative samples during the training. This aspect is critical for training the region proposals on-line. In fact, as explained above, the set of negative proposals is usually very redundant and large, since it is un-selected at this stage. Hence, for the on-line region proposal training, the Miniboostrap is key to reduce the computation and memory requirements. %Moreover, we further exploit the sub-sampling happening in the Minibootstrap, to significantly reduce the computation required for the feature extraction, anticipating the positives selection and negatives sub-sampling. This allows us to compute and use only the features for the selected regions, saving time and memory. 
%(more details about the memory saving are provided in Sec.~\ref{experiments:benchmark}).\\
Finally, the positive region proposals are refined. Similarly to the classification case, the region refinement in the original RPN is performed by a convolutional layer with $A \times 4$ channels with kernel 1 and stride 1. In the \textit{On-line RPN}, we substitute them with $A \times 4$ RLS regressors~\cite{girshick2014_rcnn}.

%%%%%%%%%%%%%%%%%%%%%%%%%%%%%%%%%%%%%%%%%%%%%%%%%%%%%%%%%%%%%%%%%%%%%%%%%%%%%%%%
\section{EXPERIMENTS}
\label{sec:experiments}
%!TEX root = ../root.tex
In this section, we report on the empirical analysis performed to demonstrate the effectiveness of the presented approach. In Sec.~\ref{experiments:setup}, we describe the experimental setup. In Sec.~\ref{experiments:preliminary}, we show that the RPN adaptation to the task at hand increases the performance of an object detection system. Then in Sec.~\ref{experiments:benchmark}, we benchmark the proposed pipeline on two robotic datasets, proving its effectiveness. In Sec.~\ref{experiments:ablation} we analyze the proposed method, showing different speed/accuracy trade-offs. Finally, in Sec.~\ref{experiments:test}, we analyze the hyper-parameters selection to show that the proposed hybrid pipelines are more robust against over-fitting than their deep-learning based counterparts.

\begin{figure*}
    %\vspace*{10mm}
	\centering
	\includegraphics[width=0.7\linewidth]{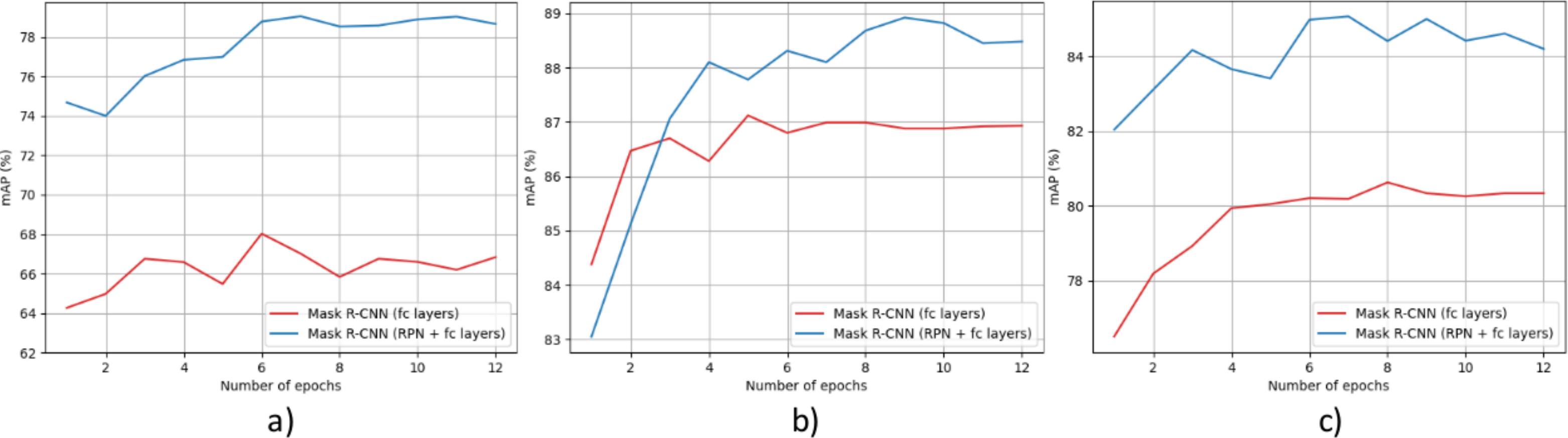}
	\caption{mAP trends for growing number of training epochs. The three plots compare the performance of Mask R-CNN (fc layers) with Mask R-CNN (RPN + fc layers). The considered TARGET-TASK is TABLE-TOP, while performance are reported for three different FEATURE-TASKs: a) 100 objects from the iCWT~\cite{pasquale2019}, b) MS COCO~\cite{coco} and c) Pascal VOC~\cite{pascal2010}.}
	\label{fig:preliminary}
\end{figure*}

\subsection{Experimental setup}
\label{experiments:setup}
In our experiments, we compare to Mask R-CNN as baseline. For a fair comparison, we consider the weights learned by training Mask R-CNN on the FEATURE-TASK (see Sec.~\ref{methods:overview}, off-line stage) and use them for: (i) the \textit{Feature Extraction Module} of our pipeline (\textbf{Ours}), and (ii) as a warm restart for the fine-tuning of the RPN and the last fc layers of Mask R-CNN on the TARGET-TASK (i.e. we set the learning rate to 0 for all the layers of the network except the last layers of the RPN and the fc layers) (\textbf{Mask R-CNN (RPN + fc layers)}).
Moreover, we consider as lower bounds for our approach two different methods, namely, (i) the \textbf{O-OD} and (ii) the \textbf{Mask R-CNN (fc layers)}. In both cases, the feature extractor is trained on the FEATURE-TASK and respectively, in the first case, the \textit{On-line Detection Module} is trained on the TARGET-TASK (\textbf{O-OD}), while in the second case, the last fc layers are optimized (\textbf{Mask R-CNN (fc layers)}).
Finally, we complete the analysis by reporting in our experiments the upper bound of our method, obtained by training the full architecture of Mask R-CNN on the TARGET-TASK (\textbf{Mask R-CNN (full)}).
We consider ResNet50~\cite{He2015} as CNN backbone for feature extraction, integrated within the Mask R-CNN as suggested in~\cite{He2015}.
We set the number of epochs for the learning of the CNN baselines by choosing the model at the epoch achieving the highest mAP on a validation set, while we select on the same validation set the main hyper-parameters of FALKON (the Gaussian kernel's variance $\sigma$ and the regularization parameter $\lambda$). We use the official Python implementation of FALKON\footnote{\url{https://github.com/falkonml/falkon}}. % and for training the On-line RPN we adopt the Logistic Loss.\\
We demonstrate the effectiveness of the proposed approach by considering the following metrics: the Average Recall \textbf{(AR (\%))} of the predicted region proposals, as defined in~\cite{hosang2016}; the mean Average Precision (\textbf{mAP (\%)}) of the final detection predictions, as defined in~\cite{pascal2010} and the time required to optimize the methods on the TARGET-TASK (\textbf{Training Time}). In particular, for the Mask R-CNN baselines, the training time is the time required to train or fine-tune Mask R-CNN via backpropagation. For \textbf{O-OD} and \textbf{Ours}, instead, it is composed of the feature extraction times, which in practical robotic applications can be absorbed (completely for \textbf{O-OD} and partially \textbf{Ours}) during the data acquisition phase and of the training time of the on-line modules\footnote{\label{video_oor}\url{https://youtu.be/HdmDYIL48H4}}\textsuperscript{,}\footnote{\label{video_ood}\url{https://youtu.be/eT-2v6-xoSs}}.
All experiments reported in this paper have been performed on a machine equipped with Intel(R) Xeon(R) E5-2690 v4 CPUs @2.60GHz, and a single NVIDIA(R) Tesla P100 GPU.

\noindent{{\bf Datasets.}}
%We considered two robotic datasets for the proposed empirical evaluation. The first dataset is the iCubWorld Transformations~\cite{pasquale2019} (referred to as iCWT) which contains images for 200 objects belonging to 20 different categories (10 instances for each category). Each object is acquired in two separate days with different sequences representing specific viewpoint transformations: 2D rotation (2D ROT), generic rotation (3D ROT), translation (TRANSL), scaling (SCALE) and a sequence that contains all transformations randomly combined (MIX). Please, refer to~\cite{pasquale2019} for further details about the dataset.
%The second dataset is a collection of table-top sequences~\cite{maiettini2019b} (referred to as TABLE-TOP) depicting 21 of the 200 objects from the iCWT, randomly placed on a table with two different table cloths: (i) pink/white pois (POIS) and (ii) white (WHITE). Refer to~\cite{maiettini2019b} for further details about the dataset. Both datasets are publicly available\footnote{\url{https://robotology.github.io/iCubWorld/\#icubworld-transformations-modal/}}. Finally, for the preliminary analysis we also consider the computer vision datasets MS COCO~\cite{coco} and Pascal VOC~\cite{pascal2010}.
For the proposed empirical evaluation, we considered two robotic datasets that depict the HRI scenario targeted in our work, showing that updating the RPN is particularly effective when the FEATURE-TASK and the TARGET-TASK represent different domains. \\
The first dataset is the iCubWorld Transformations~\cite{pasquale2019} (referred to as iCWT) which contains images for 200 objects belonging to 20 different categories (10 instances for each category). Each object is acquired in two separate days with different sequences representing specific viewpoint transformations: 2D rotation (2D ROT), generic rotation (3D ROT), translation (TRANSL), scaling (SCALE) and a sequence that contains all transformations randomly combined (MIX). Please, refer to~\cite{pasquale2019} for further details about the dataset. Since our work is an evolution of \cite{maiettini2019a}, providing results on iCWT allows to directly compare with the baseline. 
The second dataset is a collection of table-top sequences~\cite{maiettini2019b} (referred to as TABLE-TOP) depicting 21 of the 200 objects from the iCWT, randomly placed on a table with two different table cloths: (i) pink/white pois (POIS) and (ii) white (WHITE). Refer to~\cite{maiettini2019b} for further details about the dataset. Both datasets are publicly available\footnote{\url{https://robotology.github.io/iCubWorld/\#icubworld-transformations-modal/}}. \\
Finally, for the preliminary analysis described in Sec. \ref{experiments:preliminary} which gave us the motivation to pursue this work,  we also considered the computer vision datasets MS COCO~\cite{coco} and Pascal VOC~\cite{pascal2010}.

\subsection{General Vs. Updated RPN}
\label{experiments:preliminary}
Key for the fast adaptation of the learning model proposed in~\cite{maiettini2019a} is the decoupling of the optimization of the \textit{Feature Extraction Module} and of the \textit{On-line Detection Module}. Indeed, the former is trained off-line only once on a FEATURE-TASK, while the latter can be trained on-line on the TARGET-TASK. This allows the fast adaptation of the architecture to novel tasks, however it was also observed that the lack of optimization of the feature extractor might produce in some cases a performance drop~\cite{maiettini2019a}. We show in this section that this gap can be partially or completely recovered by training the region proposal on the TARGET-TASK. To prove this statement, we compare three baselines with increasing levels of adaptation of Mask R-CNN's network on the TARGET-TASK. Specifically, we considered: (i) \textbf{Mask R-CNN (fc layers)}, (ii) \textbf{Mask R-CNN (RPN + fc layers)} and (iii) \textbf{Mask R-CNN (full)} (see Sec.~\ref{experiments:setup}).\\
We considered as TARGET-TASK the TABLE-TOP dataset (see Sec.~\ref{experiments:setup}). Specifically, we take the WHITE sequence as training set and the POIS as test set. As upper bound we trained Mask R-CNN on the TARGET-TASK, obtaining an mAP on the test set of 88.2\%. For \textbf{Mask R-CNN (fc layers)} and \textbf{Mask R-CNN (RPN + fc layers)}, we tried three different FEATURE-TASKs: (i) a 100 objects identification task from the iCWT (chosen by excluding the ones contained in the TARGET-TASK) (see Fig.~\ref{fig:preliminary}a), the MS COCO dataset~\cite{coco} (see Fig.~\ref{fig:preliminary}b) and the Pascal VOC dataset~\cite{pascal2010} (see Fig.~\ref{fig:preliminary}c).
As it can be noticed in Fig.~\ref{fig:preliminary}, in all cases, there is an accuracy gap that comes from the fact that the feature extractor is not adapted on the TARGET-TASK. We note that, in the case of COCO, the gap is relatively smaller because the FEATURE-TASK contains some of the object categories of the TARGET-TASK. Importantly, this gap is partly or completely recovered when the RPN is also optimized. This result motivated our interest in implementing a fast learning method for the RPN.

\subsection{Benchmark on robotic settings}
\label{experiments:benchmark}
We validate the proposed fast learning approach on two specific robotic scenarios. In both cases, we considered as FEATURE-TASK the same 100 objects identification task from the iCWT described in Sec.~\ref{experiments:preliminary}, while varying the setting of the considered TARGET-TASK.\\

\begin{table*}[]
    %\vspace*{10mm}
	\centering
	\begin{tabular}{c|c|c|c|c}
		\cline{1-5}
		\centering \textbf{Method}                                             &\textbf{Dataset}    & \textbf{mAP  (\%) }    & \textbf{AR (\%)}             & \textbf{Train Time} \\ \hline
		\multicolumn{1}{c|}{Mask R-CNN~\cite{He2017} (full)}   				   & iCWT               &        74.2            & 46.0                         & 3h 42m               \\ \hline
		\multicolumn{1}{c|}{Mask R-CNN~\cite{He2017} (fc layers)}  		       & iCWT               &        \textbf{75.1}   & 45.8                         & 1h 23m               \\  
		\multicolumn{1}{c|}{O-OD}   	                                       & iCWT               &        73.2            & 45.8                         & \textbf{9m 13s}              \\ \hline
		\multicolumn{1}{c|}{Mask R-CNN~\cite{He2017} (RPN + fc layers)}        & iCWT               &        77.5            & 50.5                         & 2h 5m                \\ 
		\rowcolor[HTML]{A8E4A0} \multicolumn{1}{c|}{Ours}                      & iCWT               &        \textbf{80.3}   & \textbf{53.0}                 & \textbf{23m 11s}              \\ \hline
		\multicolumn{1}{c|}{Mask R-CNN~\cite{He2017} (full)}   				   & TABLE-TOP          &        87.2            & 55.4                         & 18m 49s               \\ \hline
		\multicolumn{1}{c|}{Mask R-CNN~\cite{He2017} (fc layers)}  		       & TABLE-TOP          &        67.0            & 35.7                         & 12m 50s               \\  
		\multicolumn{1}{c|}{O-OD}   	                                       & TABLE-TOP          &        \textbf{73.0}   & 35.7                         & \textbf{2m 19s}              \\ \hline
		\multicolumn{1}{c|}{Mask R-CNN~\cite{He2017} (RPN + fc layers)}        & TABLE-TOP          &        78.6            & 50.4                         & 28m 2s                \\ 
		\rowcolor[HTML]{A8E4A0} \multicolumn{1}{c|}{Ours}                      & TABLE-TOP          &        \textbf{82.4}   & \textbf{51.6}                & \textbf{5m 22s}              \\ \hline
	\end{tabular}
	%\caption{Benchmark on \textsc{iCWT} and \textsc{TABLE-TOP}. The compared methods are: (i) the full training of Mask R-CNN \textbf{(Mask R-CNN (full))}, (ii) the fine-tuning of the last fc layers of Mask R-CNN (\textbf{Mask R-CNN (fc layers)}), (iii) the On-line Object Detection (\textbf{O-OD}), (iv) the fine-tuning of both the RPN and last fc layers (\textbf{Mask R-CNN (RPN + fc layers)}) and (v) the proposed approach (\textbf{Ours}).}
	\caption{Benchmark on \textsc{iCWT} and \textsc{TABLE-TOP}. The compared methods are: the full training of Mask R-CNN \textbf{(Mask R-CNN (full))}, the fine-tuning of the last fc layers of Mask R-CNN (\textbf{Mask R-CNN (fc layers)}), the On-line Object Detection (\textbf{O-OD}), the fine-tuning of both the RPN and last fc layers (\textbf{Mask R-CNN (RPN + fc layers)}) and the proposed approach (\textbf{Ours}).}
	\label{table:bench_validation}
\end{table*}
\noindent{{\bf iCWT: different objects, same setting.}}
%In this experiment, we considered as TARGET-TASK a 30 objects identification task from the iCWT (chosen by excluding the ones contained in the FEATURE-TASK). We collect a set of $\sim$8k images for the training set from the 2D ROT, 3D ROT, TRANSL and SCALE transformations of each object of the task and a set of 4.5k images manually annotated from the MIX sequence of each object for the test set. By following the indications provided in~\cite{maiettini2019a} for the Minibootstrap {hyper-}parameters, we set the number of iterations to 10 and the size of batches to 2000, for both the On-line RPN in \textbf{Ours} and the \textit{On-line Detection Module} in \textbf{Ours} and \textbf{O-OD}.\\
In this experiment, we considered as TARGET-TASK a 30 objects identification task from the iCWT (chosen by excluding the ones contained in the FEATURE-TASK). We collect two sets of $\sim$8k and $\sim$4.5k images automatically annotated~\cite{pasquale2019} for the training and validation sets which are intrinsically noisy, respectively, from the 2D ROT, 3D ROT, TRANSL and SCALE transformations of each object of the task and a set of 4.5k images manually annotated from the MIX sequence of each object for the test set. By following the indications provided in~\cite{maiettini2019a} for the Minibootstrap {hyper-}parameters, we set the number of iterations to 10 and the size of batches to 2000, for both the On-line RPN in \textbf{Ours} and the \textit{On-line Detection Module} in \textbf{Ours} and \textbf{O-OD}.\\
The results reported in Tab.~\ref{table:bench_validation} show that the adaptation of the region proposal allows to improve the AR of the candidate RoIs and the mAP of the detections for both the baseline (\textbf{Mask R-CNN (RPN + fc layers)}) and the proposed approach (\textbf{Ours}). Moreover, while being comparable in terms of accuracy with the fine-tuning based counterparts, the on-line approaches (namely, in Tab.~\ref{table:bench_validation}, \textbf{O-OD} and \textbf{Ours}, in rows 3 and 5, respectively) are remarkably faster.
%, allowing to train the detection model in a fraction of the time required by the baselines. 
%Finally, 
Furthermore, Tab.~\ref{table:bench_validation} shows that the proposed approach allows to obtain the most accurate detections with an mAP of 80.3\% and an AR of 53.0\% with a training time of $\sim$23 minutes, while the baseline requires $\sim$2 hours to achieve a comparable performance.\\
Notably, in this case, the \textbf{Mask R-CNN (full)} achieves a relatively low accuracy, with respect to the other baselines. 
%This may be due to an extreme fit of the model on the intrinsically noisy annotations of the automatically collected training and validation sets. Indeed, in the next section, where manual annotations are used, the \textbf{Mask R-CNN (full)} achieves the best accuracy.
We comment and give reasons to this in Sec.~\ref{experiments:test}.\\
%As described in Sec.~\ref{methods:overview}, we chose the training hyper-parameters on the validation set, which in this experiment is intrinsically noisy (as the training set). This led the full training of Mask R-CNN \textbf{(Mask R-CNN (full))} to have one of the lowest mAPs. Nevertheless, without looking at the test set at training time, such bias can not be removed.\\
Finally, the accuracy achieved by the \textbf{O-OD} (73,2\%) allows a direct comparison with the mAP (71,2\%) reported in~\cite{maiettini2019a} (specifically, the line \textbf{FALKON + Minibootstrap 10 × 2000} in Tab. 2). Since the only difference between the two approaches is the feature extractor (i.e. Faster R-CNN in~\cite{maiettini2019a} and Mask R-CNN in this work) the difference in mAP empirically supports that employing Mask R-CNN's first layers improves the quality of the extracted features, with respect to the Faster R-CNN counterpart.\\

\noindent{{\bf TABLE-TOP: different objects and setting.}}
%In the previous experiment, we considered a TARGET-TASK different from the FEATURE-TASK in terms of classes, but similar in terms of setting (i.e. viewpoint and background). In this experiment, we further assess the fast domain adaptation capability of the proposed approach by considering as TARGET-TASK the TABLE-TOP dataset (see Sec.~\ref{experiments:setup}). Specifically, as in Sec.~\ref{experiments:preliminary}, we take the WHITE sequence as training set (2k images) and the POIS as test set (1k images). We set empirically the number of Minibootstrap's iterations and batch sizes respectively to 2 and 2000 for both the \textit{On-line RPN} in \textbf{Ours} and the \textit{On-line Detection Module} in \textbf{Ours} and \textbf{O-OD}.\\
In the previous experiment, we considered a TARGET-TASK different from the FEATURE-TASK in terms of classes, but similar in terms of setting (i.e. viewpoint and background). In this experiment, we further assess the fast domain adaptation capability of the proposed approach by considering as TARGET-TASK the TABLE-TOP dataset (see Sec.~\ref{experiments:setup}). Specifically, as in Sec.~\ref{experiments:preliminary}, we take 2k and 500 images from the WHITE sequence as training and validation sets, respectively, and 1k imaged from the POIS sequence as test set. As described in~\cite{maiettini2019b} all these images are manually annotated. We set empirically the number of Minibootstrap's iterations and batch sizes respectively to 2 and 2000 for both the \textit{On-line RPN} in \textbf{Ours} and the \textit{On-line Detection Module} in \textbf{Ours} and \textbf{O-OD}. 
Please, note that while \textbf{Mask R-CNN (full)} has been trained for 4 epochs, \textbf{Mask R-CNN (RPN + fc layers)}) required 9 epochs to be optimized, with the aforementioned validation protocol. Further comments on this are reported in Sec.~\ref{experiments:test}.
%This explains why \textbf{Mask R-CNN (full)}'s training time is lower than \textbf{Mask R-CNN (RPN + fc layers)}'s. \federico{The analysis on the test in Sec.~\ref{experiments:test} addresses this behavior.} 

As it can be inferred from the last rows of Tab.~\ref{table:bench_validation}, in this case, the greater difference in the setting between the FEATURE-TASK and the TARGET-TASK leads to a larger gap in terms of AR and mAP between the upper bound (\textbf{Mask R-CNN (full)}) and the \textbf{O-OD} and the fine-tuning of the last fc layers. The adaptation of the RPN allows to partially recover this gap for the baseline and the \textit{On-line RPN} of the proposed approach on the TARGET-TASK. However, while the first one (\textbf{Mask R-CNN (RPN + fc layers)}) requires $\sim$28 minutes of training, the proposed detection method (\textbf{Ours}) can be optimized in $\sim$5 minutes, achieving the best mAP (82.4\%) and AR (51.6\%) between the two.\\
% Finally, we would like to recall that, according to the training protocol described in Sec.~\ref{methods:overview}, the choice of the number of the training epochs of the baselines (i.e. \textbf{Mask R-CNN (full)}, \textbf{Mask R-CNN (fc layers)} and \textbf{Mask R-CNN (RPN + fc layers)}) is made through the validation set, this explains why \textbf{Mask R-CNN (full)}'s training time is lower than the time required for training \textbf{Mask R-CNN (RPN + fc layers)}.\\
\begin{figure}
    %\vspace*{5mm}
	\centering
	\includegraphics[width=0.7\linewidth]{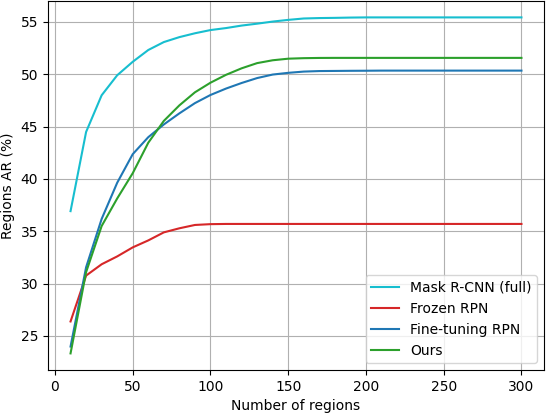}
	\caption{AR trends for varying number of predicted RoIs. The methods compared are: (i) the RPN learned within the full training process of Mask R-CNN (\textbf{Mask R-CNN (full)}), (ii) the RPN trained on the FEATURE-TASK (\textbf{Frozen RPN}), (iii) the RPN trained within the \textbf{Mask R-CNN (RPN + fc layers)} baseline (\textbf{Fine-tuning RPN}) and (iv) the proposed approach (\textbf{Ours}). }
	\label{fig:regrec}
\end{figure}

\begin{table*}[]
     %\vspace*{10mm}
 	\centering
 	\begin{tabular}{c|c|c|c|c}
 		\cline{1-5}
 		\centering \textbf{Method}                                             &\textbf{Dataset}    & \textbf{mAP  (\%) on Test Set}     & \textbf{ \#Epochs}                & \textbf{Avg. Train Time per Epoch}  \\ \hline
 		\multicolumn{1}{c|}{Mask R-CNN~\cite{He2017} (full)}   				   & iCWT               &        79.8                            &  3                                       & 24m 35s                          \\ \hline
 		\multicolumn{1}{c|}{Mask R-CNN~\cite{He2017} (fc layers)}  		       & iCWT               &        78.9                            &  1                                       & 7m 36s                           \\  
 		\multicolumn{1}{c|}{O-OD}   	                                       & iCWT               &        73.2                            &  -                                       &  -                                    \\ \hline
 		\multicolumn{1}{c|}{Mask R-CNN~\cite{He2017} (RPN + fc layers)}        & iCWT               &        79.9                            &  1                                       & 12m 25s                          \\ 
 		\rowcolor[HTML]{A8E4A0} \multicolumn{1}{c|}{Ours}                      & iCWT               &        80.3                            &  -                                       &  -                           \\ \hline
 		\multicolumn{1}{c|}{Mask R-CNN~\cite{He2017} (full)}   				   & TABLE-TOP          &        88.2                            &  5                                       & 4m 27s                                   \\ \hline
		\multicolumn{1}{c|}{Mask R-CNN~\cite{He2017} (fc layers)}  		       & TABLE-TOP          &        68.0                            &  6                                       & 1m 50s                                   \\  
 		\multicolumn{1}{c|}{O-OD}   	                                       & TABLE-TOP          &        73.8                            &  -                                       &  -                            \\ \hline
 		\multicolumn{1}{c|}{Mask R-CNN~\cite{He2017} (RPN + fc layers)}        & TABLE-TOP          &        79.1                            &  7                                       & 3m 7s                                    \\ 
		\rowcolor[HTML]{A8E4A0} \multicolumn{1}{c|}{Ours}                      & TABLE-TOP          &        84.5                            &  -                                       &  -                            \\ \hline
	\end{tabular}
 	\caption{Analysis of the performance on test set of \textsc{iCWT} and \textsc{TABLE-TOP}. The compared methods are the same as the ones of Tab.~\ref{table:bench_validation}. The number of epochs is not reported for the O-OD and Ours due to the different training protocol w.r.t. the baselines.}
	\label{table:bench_test}
\end{table*}

\subsection{Analysis of the region proposals}
\label{experiments:ablation}
In Sec.~\ref{experiments:preliminary}, we demonstrated that adapting the region proposals to the TARGET-TASK increases the detection performance. Then, in Sec.~\ref{experiments:benchmark} we showed that the proposed approach allows to do it significantly faster than the fine-tuning based counter part, while keeping or improving performance. In this section, we specifically dig into the analysis of the region proposals and show that, indeed, tuning the region proposals to the TARGET-TASK improves their quality. This effectively reduces the number of RoIs required to achieve a target recall, which, in turn, contributes to reduce the computational load of the detection task, thus achieving the same performance more efficiently.
To prove this effect, we considered again the 100 objects identification task introduced in Sec.~\ref{experiments:preliminary} as FEATURE-TASK and the TABLE-TOP dataset as TARGET-TASK. We compare in Fig.~\ref{fig:regrec} the average recall obtained with a growing number of predicted RoIs, for the following methods: (i) the proposed On-line RPN (\textbf{Ours}), (ii) the RPN learned within the full training process of Mask R-CNN (\textbf{Mask R-CNN (full)}), (iii) the RPN trained only on the FEATURE-TASK (\textbf{Frozen RPN} in Fig.~\ref{fig:regrec}) and (iv) the RPN trained within the \textbf{Mask R-CNN (RPN + fc layers)} baseline (\textbf{Fine-tuning RPN} in Fig.~\ref{fig:regrec}). \\
Fig.~\ref{fig:regrec} shows that in the four cases, the AR saturates with more than 150 RoIs. Moreover, while the RPN trained only on the FEATURE-TASK reasonably achieves the worst performance, the proposed approach allows to achieve an AR higher than its baseline (i.e. \textbf{Fine-tuning RPN}). Therefore, our method achieves the AR that is the closest to the upper bound (\textbf{Mask R-CNN (full)}). This allows to significantly reduce the number of proposals when training the \textit{On-line Detection Module} and at inference time, while preserving performance in terms of AR. For instance, by reducing the number of regions from 300 (as in Sec.~\ref{experiments:benchmark}, \textbf{TABLE-TOP}) to 100, we manage to reduce the computational load of the method by a factor of 3, while only slightly reducing the obtained AR (see Fig.~\ref{fig:regrec}). Finally, this trade-off can be brought to the extreme, by considering 50 regions. In this case, the computational load is further reduced, while the AR is still higher than the baseline (\textbf{Frozen RPN}).

%This allows to significantly reduce the number of proposals when training the \textit{On-line Detection Module} and at inference time, while preserving detection performance. For instance, by reducing the number of regions from 300 (as in Sec.~\ref{experiments:benchmark}, Table-top) to 50, we manage to decrease the training time to $\sim$5 minutes (instead of $\sim$6 minutes as in Tab.~\ref{table:ttop}), while achieving an mAP of 79.4\%, which is still higher than the performance obtained by \textbf{O-OD} (70.5\%) and the baseline \textbf{Mask R-CNN (RPN + fc layers)} (78.6\%).

%adattare la region proposal method sul target task ha l'evidente beneficio che l'accuracy migliora, however c'è anche il beneficio che si riduce il numero delle regioni necessarie sia a train che ad inference time. Questo succede perchè le regioni proposte da una RPN addestrata sono più precise e quindi ne occorrono di meno.
%\noindent{{\bf Minibootstrap parameters variation}}
%
%\begin{enumerate}
%	\item show different speed-accuracy trade off
%\end{enumerate}

%\noindent{{\bf Regions recall analysis}}
%\begin{enumerate}
%	\item Recall variation for different numbers of region proposals (the purpose is to show that by retraining the rpn online we are able to decrease the number of necessary regions to do training and inference, reducing the inference/training time)
%\end{enumerate}

%\noindent{{\bf Logistc loss vs L2 loss (?)}}
\subsection{Analysis of hyper-parameters selection}
\label{experiments:test}
 All results reported in Sec.~\ref{experiments:benchmark} have been achieved by choosing the hyper-parameters on a validation set as described in Sec.~\ref{experiments:setup}. This is common practice both in computer vision and robotics, because the test set is not available at training time and it is especially important in on-line robotic applications, where hyper-parameters should be set off-line, before model's deployment.
 However, even though in practical cases this is the only available choice, tuning parameters on the validation set (which is typically drawn from the training set) might not lead to the optimal values for the test set.
 To analyze this phenomenon for the experiments reported in Sec.~\ref{experiments:benchmark}, we report in Tab.~\ref{table:bench_test} the best mAP values on the test sets of iCWT and TABLE-TOP for all the pipelines compared in Sec.~\ref{experiments:benchmark}. These are obtained by optimizing the hyper-parameters on the test set. Note that this analysis is useful to understand the aforementioned phenomenon, however practically, this cannot be done since test data is not available at training time.
%As it can be observed, all the mAP values are higher than the ones reported in Tab.~\ref{table:bench_validation}. More importantly, it can be noticed that the pipelines that are the most affected are the Mask R-CNN baselines. For instance, the best accuracy achieved by \textbf{Mask R-CNN (full)} on iCWT is 79.8 (see table Tab.~\ref{table:bench_test}) while the value obtained by choosing hyper-parameters with the validation set is 74.2 (see table Tab.~\ref{table:bench_validation}). Finally, in some cases Mask R-CNN baselines achieve the best accuracy on the test set with less epochs of training, also reducing training time. However, note that, these shorter training times cannot be achieved in practice. First of all, one reason is that we cannot tune the hyper-parameters on the test set, secondly an additional time for data acquisition on the robot needs to be summed to the ones reported in Tabs.~\ref{table:bench_validation} and ~\ref{table:bench_test} (this time is already counted in the training time of \textbf{O-OD} and \textbf{Ours} since it can be absorbed by the feature extraction).
As it can be observed, almost all the mAP values are higher than the ones reported in Tab.~\ref{table:bench_validation}. More importantly, it can be noticed that the pipelines that are the most affected are the Mask R-CNN baselines. For instance, the best accuracy achieved by \textbf{Mask R-CNN (full)} on iCWT is 79.8\% (see Tab.~\ref{table:bench_test}) while the value obtained by choosing hyper-parameters with the validation set is 74.2\% (see Tab.~\ref{table:bench_validation}). Finally, in some cases Mask R-CNN baselines achieve the best accuracy on the test set with less epochs of training, also reducing training time. However, note that, these shorter training times cannot be achieved in practice. Firstly, the hyper-parameters can not be tuned on the test set. Secondly, an additional time for data acquisition on the robot needs to be summed to the ones reported in Tabs.~\ref{table:bench_validation} and ~\ref{table:bench_test}, since the training of Mask R-CNN baselines require the availability of all the training images to start, for instance for data shuffling to compose the training batches. Instead, for \textbf{O-OD} and \textbf{Ours}, this time can be absorbed during the feature extraction phase and it is already counted in the training time (see \ref{experiments:setup}). 

Finally, in Tab.~\ref{table:bench_test}, it can be noted that \textbf{O-OD} and \textbf{Ours} do not present the aforementioned accuracy and time gaps (note that, due to their training protocol training time remains constant) which occur for the Mask R-CNN baselines. This represents a major advantage of the proposed approach which turns out to be less prone to overfit the training set, allowing to choose close-to-optimal hyper-parameters off-line.
%\subsection{Analysis on the test set}
%\label{experiments:test}

%%%%%%%%%%%%%%%%%%%%%%%%%%%%%%%%%%%%%%%%%%%%%%%%%%%%%%%%%%%%%%%%%%%%%%%%%%%%%%%%
\section{CONCLUSIONS}
\label{sec:conclusions}
%!TEX root = ../root.tex
High reliability and fast adaptation to new tasks are two equally important requirements for robotic vision systems. While current deep learning methods achieve remarkable performance in terms of accuracy, their training time is typically high: this prevents their adoption in those robotic applications which require adaptation in presence of new scenarios or novel objects. In this paper, we present an approach for %quickly training an object detector
quickly adapting an object detector to a new task. We overcome a performance gap in previous work~\cite{maiettini2019a}, by proposing a novel pipeline which adapts both the region proposal and the classification layers as new data comes.  We demonstrate, through a set of experiments on different datasets, the effectiveness of the proposed approach. 
For these reasons, we believe this work makes an important step to achieve more flexible robotic vision systems.

%\clearpage
% The acknowledgments are automatically included only in the final version of the paper.
%\acknowledgments{Acks}

%===============================================================================

% no \bibliographystyle is required, since the corl style is automatically used.
\bibliographystyle{unsrt}
\bibliography{bibliography}  % .bib

\end{document}